\documentclass[10pt,twocolumn,letterpaper]{article}

\usepackage{cvpr}
\usepackage{times}
\usepackage{epsfig}
\usepackage{graphicx}
\usepackage{amsmath}
\usepackage{amssymb}
\usepackage{multirow}
\usepackage[utf8]{inputenc}
\usepackage{enumitem}
\usepackage{amssymb}
\usepackage{pifont}
\newcommand{\cmark}{\ding{51}}%
\newcommand{\xmark}{\ding{55}}%

\usepackage[pagebackref=true,breaklinks=true,letterpaper=true,colorlinks,bookmarks=false]{hyperref}

\begin{document}

\title{DeePhy: On Deepfake Phylogeny} 

\author{Kartik Narayan, Harsh Agarwal, Kartik Thakral, Surbhi Mittal, Mayank Vatsa, Richa Singh
\and
IIT Jodhpur, India\\
{\tt\small \{narayan.2, agarwal.10, thakral.1, mittal.5, mvatsa, richa\}@iitj.ac.in}
}
\maketitle
\thispagestyle{empty}

\begin{abstract}
   Deepfake refers to tailored and synthetically generated videos which are now prevalent and spreading on a large scale, threatening the trustworthiness of the information available online. While existing datasets contain different kinds of deepfakes which vary in their generation technique, they do not consider progression of deepfakes in a ``phylogenetic'' manner. It is possible that an existing deepfake face is swapped with another face. This process of face swapping can be performed multiple times and the resultant deepfake can be evolved to confuse the deepfake detection algorithms. Further, many databases do not provide the employed generative model as target labels. Model attribution helps in enhancing the explainability of the detection results by providing information on the generative model employed. In order to enable the research community to address these questions, this paper proposes DeePhy, a novel Deepfake Phylogeny dataset which consists of 5040 deepfake videos generated using three different generation techniques. There are 840 videos of one-time swapped deepfakes, 2520 videos of two-times swapped deepfakes and 1680 videos of three-times swapped deepfakes. With over 30 GBs in size, the database is prepared in over 1100 hours using 18 GPUs of 1,352 GB cumulative memory. We also present the benchmark on DeePhy dataset using six deepfake detection algorithms. The results highlight the need to evolve the research of model attribution of deepfakes and generalize the process over a variety of deepfake generation techniques. The database is available at: \href{http://iab-rubric.org/deephy-database}{http://iab-rubric.org/deephy-database}
\end{abstract}
\begin{figure}[t] 
\centering
\includegraphics[scale=0.47]{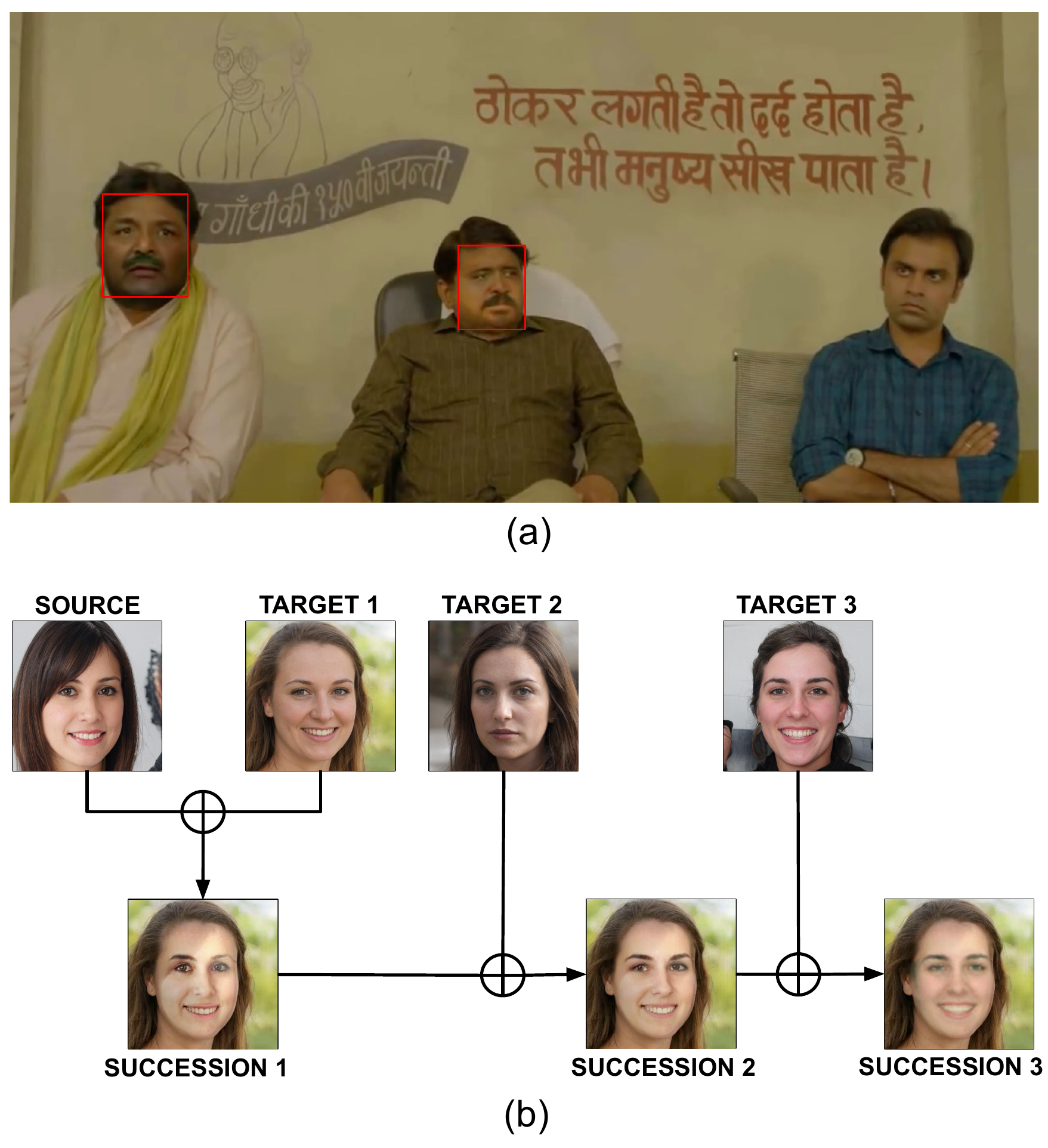}
\caption{(a) Sample of multi-face deepfakes in a single frame. Subject-1 and Subject-2 (in red-bounding boxes) are deepfakes generated using two different algorithms. Subject-3 is a real face. (b) Pictorial representation of the evolution of deepfakes from the same source identity with different targets in a phylogenetic manner. \textit{Deepfake Phylogeny} refers to the idea of generation of deepfakes using multiple generation techniques in a sequential manner.}
\label{fig:vabstract}
\end{figure}


\section{Introduction}
\label{introduction}
``Deepfake" is an amalgamation of the words ``deep" and ``fake" which implies generating fake multimedia content like images, videos or audio through deep learning techniques. 
Misinformation in the form of deepfakes is detrimental to the society and hampers the credibility of information being shared on the internet. In order to address this issue, a number of datasets such as FF++~\cite{ff++}, DFDC~\cite{dfdc}, and Celeb-DF~\cite{celebdf} have been proposed in the past that have accelerated the development of deepfake detection. Many of these are either generated using a single generative technique~\cite{deepfaketimit, celebdf, FaceForensics, uadfv} or do not provide attribution of the generation models employed~\cite{wilddf}. It can be anticipated that more robust and complicated deepfake generation techniques would be introduced in the future and to handle this imminent threat; detection methodologies must have a thorough comprehension and development. While existing detection methods demonstrate promising results, categorizing an image as fake fails to provide any information pertaining to the generative technique utilised and many questions still remain unanswered. The amalgamation of model attribution along with deepfake detection can be utilised as a way to track the history of an image. 
 In recent work from Le et al.~\cite{openforensics}, the concept of multi-face deepfakes is introduced. This concept emphasises on the use of deepfake generation techniques for generating more than one deepfakes in a single frame. It means identifying real faces in a frame and swapping them with other real face images to generate a deepfake as can be seen in Figure \ref{fig:vabstract}(a). It is evident that there is a growing demand of deepfakes and the field of deepfakes is now becoming multifarious. Inspired from the research in the field of phylogeny~\cite{ phylo4, phylo1, phylo2, phylo3, phylo5} in various domains, we introduce the concept of Deepfake Phylogeny which refers to usage of multiple generative models sequentially to generate deepfake videos as illustrated in Figure \ref{fig:vabstract}(b). It brings forth the idea of evolution of deepfakes and opens up new challenges like model attribution and prediction of the sequential order of deepfake techniques employed in the phylogeny.

In this paper, we focus on the problem of model attribution in case of Deepfake Phylogeny.Pre-trained networks are being widely employed professionally for artistic works, content creation or for entertainment, their creators must be aware about its usage and must be duly credited to avoid intellectual property theft and illegal use of deepfake technology~\cite{attributing_fake}. Model attribution makes the inferred results more trustworthy and enhances the associated confidence as opposed to categorization by black box deepfake detection algorithms~\cite{9105991}. The problem of model attribution has been recently explored \cite{faceswap_attribute, attribute_df}; however, it has largely been based on synthetic videos generated using a single generative technique. It is possible that a deepfake video is generated using multiple methods that are applied sequentially on a video. In real-world scenarios, an adversary can employ fake videos generated using one technique and use another publicly available model to further manipulate the video. In these cases, single technique attribution models might not work as they may not have seen such samples during training. Model attribution helps in holding accountable the original actor in the video. 
If the generative technique is known, a decoder network can be employed for reverse-engineering the source actor of the deepfake video~\cite{meta}. This is of great use for the law-enforcement authorities to identify repetitive offenders spreading misinformation and thereby maintain a vigil over their activities. The same can be extended for phylogenetic videos by considering the sequence of generative techniques.

With the above motivation, we propose \textit{DeePhy}, a novel Deepfake Phylogeny dataset comprising of 5040 videos. The videos are generated using 100 unique videos containing subjects of Indian origin sourced from \textit{YouTube}. For generating these manipulated videos, FaceShifter~\cite{faceshifter}, FSGAN~\cite{nirkin2019fsgan}, and FaceSwap~\cite{deepfakes} techniques are utilized. The first two methods are GAN-based whereas the third is an Autoencoder based model. We have segregated the generated deepfakes into three main categories: (1) Using single technique, (2) Using two different techniques and (3) Using three different techniques. We formulate the problem of model attribution in Deepfake Phylogeny as a multi-label classification problem and sequence prediction as a multi-class classification problem. We predict the models used for generation irrespective of their order of utilization in the first case and in the later, we predict the sequence in which the deepfake generation techniques are employed. We provide thorough benchmark results using various popular models on the proposed DeePhy dataset.

\begin{figure*}[t] 
\centering
\includegraphics[width=\textwidth]{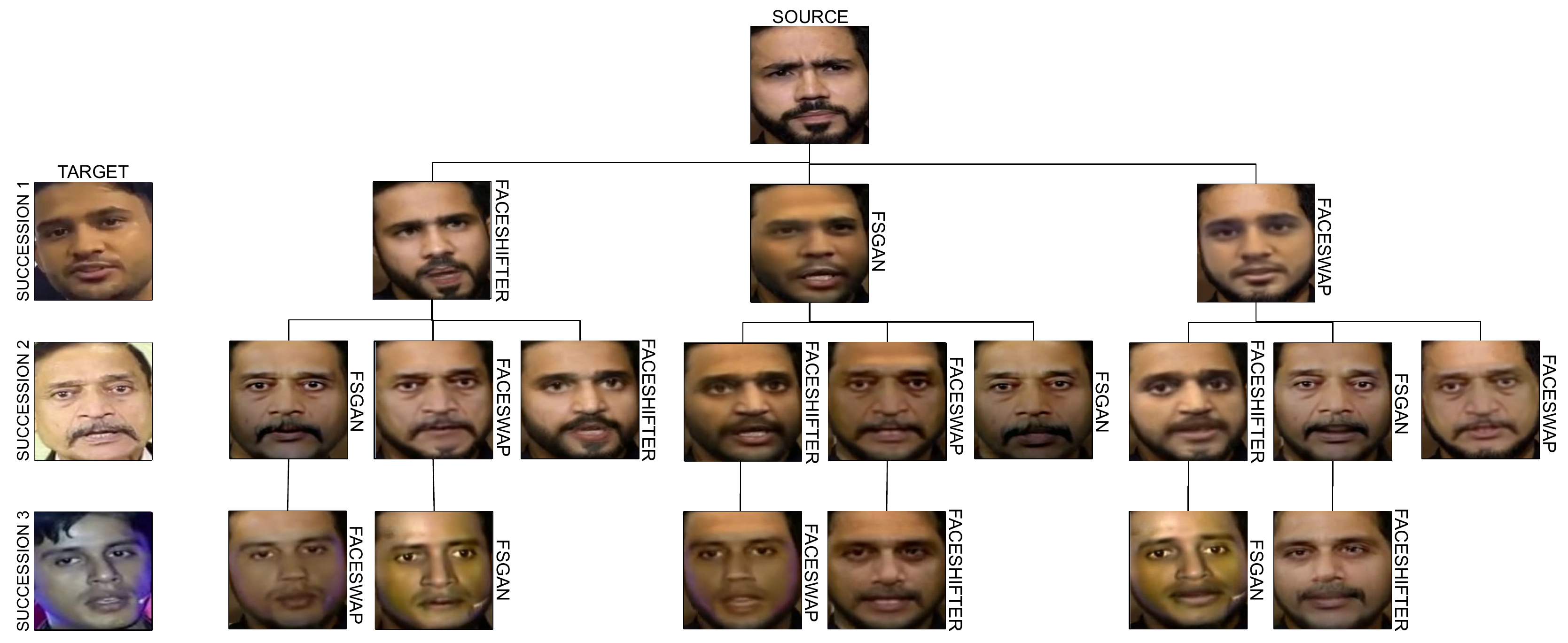}
\caption{Phylogeny tree illustrating the three deepfake generation techniques of FaceShifter, FSGAN, and FaceSwap employed successively. A single source and three targets are used at different levels of the tree.}
\label{fig:visual_abstract}
\vspace{-5pt}
\end{figure*}

\section{Related Work}
\label{sec:related}
\noindent As the number of deepfake generation techniques grow with increasing sophistication, many datasets have been proposed to build robust deepfake detection models. \\
\noindent \textbf{Deepfake Datasets:}
The necessity of spotting such digitally altered data is becoming increasingly important as the number and accessibility of deepfake generation techniques grow. While there is a significant number of such altered videos/images on the internet, benchmark datasets particularly generated for deepfake detection research are in high demand so as to evaluate novel methods of generation. Specifically, datasets can be divided on the basis of the number of generation techniques utilized into two broad categories: (1) single generation technique and (2) multiple generation techniques. Most of the datasets contain videos generated using a single, popular and accessible generation technique. Table \ref{table:related_work} summarises the the number of videos and generative techniques across different datasets.

\begin{table}[!t]
\centering
\scriptsize

\caption{Quantitative analysis of existing Deepfake datasets}
\label{table:related_work}
\begin{tabular}{|l|c|c|c|c|}
\hline
\textbf{Dataset} &  \textbf{\begin{tabular}[c]{@{}c@{}}Real\\ Videos\end{tabular}} & \textbf{\begin{tabular}[c]{@{}c@{}}Deepfake\\ Videos\end{tabular}} & \textbf{Methods} & \textbf{Phylogeny} \\ \hline
UADFV~\cite{uadfv}                                                           & 49                                                                                                            & 49                                                                                                                & 1                & \xmark                  \\ \hline
FaceForensics~\cite{FaceForensics}                                                   & 1,004                                                                                                          & 2,008                                                                                                              & 1                & \xmark                  \\ \hline
FaceForensics++~\cite{ff++}                                                 & 1,000                                                                                                          & 4,000                                                                                                              & 4                & \xmark                  \\ \hline
DFDC~\cite{dfdc}                                                            & 23,654                                                                                                         & 104,500                                                                                                            & 8                & \xmark                  \\ \hline
DeepFakeTIMIT~\cite{deepfaketimit}                                                   & 320                                                                                                           & 640                                                                                                               & 2                & \xmark                  \\ \hline
Deep Fakes Dataset~\cite{wilddf}                                              & 70                                                                                                            & 70                                                                                                                & NA               & \xmark                  \\ \hline
CelebDF~\cite{celebdf}                                                         & 590                                                                                                           & 5,639                                                                                                              & 1                & \xmark                  \\ \hline
KoDF~\cite{KoDF}                                                            & 62,166                                                                                                         & 175,766                                                                                                            & 6                & \xmark                  \\ \hline
\textbf{DeePhy}                                                   & \textbf{100}                                                                                                  & \textbf{5,040}                                                                                                     & \textbf{3}       & \textbf{\cmark}                  \\ \hline
\end{tabular}
\label{relatedwork}
\end{table}

\noindent \textbf{Deepfake-Attribution:} Ciftci et al.~\cite{umur_heartbeats} have proposed a GAN based framework that utilizes the variation in heartbeat of deepfake videos to find the source method using which the deepfake was generated. The robustness of the model was demonstrated by testing on FaceForensics++ dataset where it achieved exceptional accuracy. Jain et al.~\cite{dadhcnn} have put forward a novel network that was based on hierarchical CNNs for distinguishing between retouching and GAN produced images and identifying the particular GAN architecture. Yu et al.~\cite{attributing_fake} have proposed a novel model on the basis of an attribution network that constructs a correlation between the fingerprint and its respective image and then employs GANs to detect fake images. Zhang et al.~\cite{attribution_deepfakes} have attributed GAN generated images by employing seed reconstruction based method wherein the latent variable utilized in generator is explored by utilizing gradient descent so as to find the closest match. However, this approach cannot be applied to techniques that lack latent variable such as facial-reenactment and image-to-image translation. Jia et al.~\cite{model_attribution} have introduced a novel deepfake dataset ``Deepfakes from Different Models (DFDM)" comprising of deepfakes from five different Autoencoder based models. They also proposed a spatio-temporal attention based framework for model attribution which surpassed most of the existing deepfake detection methods. Yang et al.~\cite{yang_dfnet} have proposed a new network for deepfake network architecture attribution named DNA-Det that identifies fingerprints of GAN-based architectures by employing patchwise contrastive learning and pre-training on image transformations. Goebel et al.~\cite{GoebelGAN} have utilized XceptionNet trained on the co-occurence matrices of images for detection, attribution and localisation of GAN produced images. The authors demonstrate the performance of the aforementioned network by varying the patch-size for co-occurence matrices and applying various image quality factors (compression). Marra et al.~\cite{marra_cycle} have analysed diverse approaches for distinguishing CycleGAN images from ordinary images and achieved the best results by combining residual features~\cite{residual1, residual2} and deep learning~\cite{xceptionnet}. Except for FaceForensics++~\cite{ff++}, DeepfakeTIMIT~\cite{deepfaketimit} and KoDF~\cite{KoDF}, all other datasets are generated using a single generation technique. Also, the aforementioned datasets do not contain videos that are generated in a phylogenetic way i.e. generated using two or more techniques sequentially. 

\section{Proposed Deepfake Phylogeny Dataset}
\label{sec:proposed}
The core contribution of this work is introducing the concept of deepfake phylogeny to instigate further research with this problem statement, as illustrated in Figure~\ref{fig:visual_abstract}. For this, we propose DeePhy, a Deepfake Phylogeny dataset and benchmark the existing state-of-the-art techniques that are popular in deepfake literature \cite{afchar2018mesonet, chollet2017xception, li2018exposing, nguyen2019use, rossler2019faceforensics++}.

\subsection{DeePhy: Deepfake Phylogeny Dataset Details}
\noindent \textbf{Phylogeny Procedure:} The proposed DeePhy dataset contains videos generated using $3$ different techniques explained in detail in Section ~\ref{SynthesisMethod}. The dataset is divided into three categories namely - ``Succession 1", ``Succession 2" and ``Succession 3" with $840$, $2520$ and $1680$  videos,  respectively. Succession 1 consists of videos where the face of a target is swapped with the source once. Similarly, in Succession 2 the face of the target is swapped with two different sources and in Succession 3 the face of the target is swapped with three different sources. The technique used for swapping are not necessarily the same. Table~\ref{datasetTable} summarizes the generation techniques used in each Succession. In Succession 2 and 3, the generation techniques are applied to the previously generated deepfake in Succession 1 and 2 respectively. Table~\ref{datasetTable} describes the generation techniques used in each succession and each row of the table shows the cumulative order in which the generation techniques are used to create phylogeny. For instance, row 1 depicts that FSGAN is used for face swapping in ``Succession 1". In ``Succession 2", FaceSwap technique is used to swap the face of an existing deepfake face (generated using FSGAN) to create a phylogeny. ``Succession 3" further deepens the phylogeny by swapping face in an already phylogenetic deepfake face (generated using FSGAN followed by FaceSwap) using the FaceShifter technique. Each block in Table~\ref{datasetTable} renders a set of 280 deepfake videos and corresponding generation techniques used, along with its order.

\begin{figure}[t] 
\centering
\includegraphics[scale=0.37]{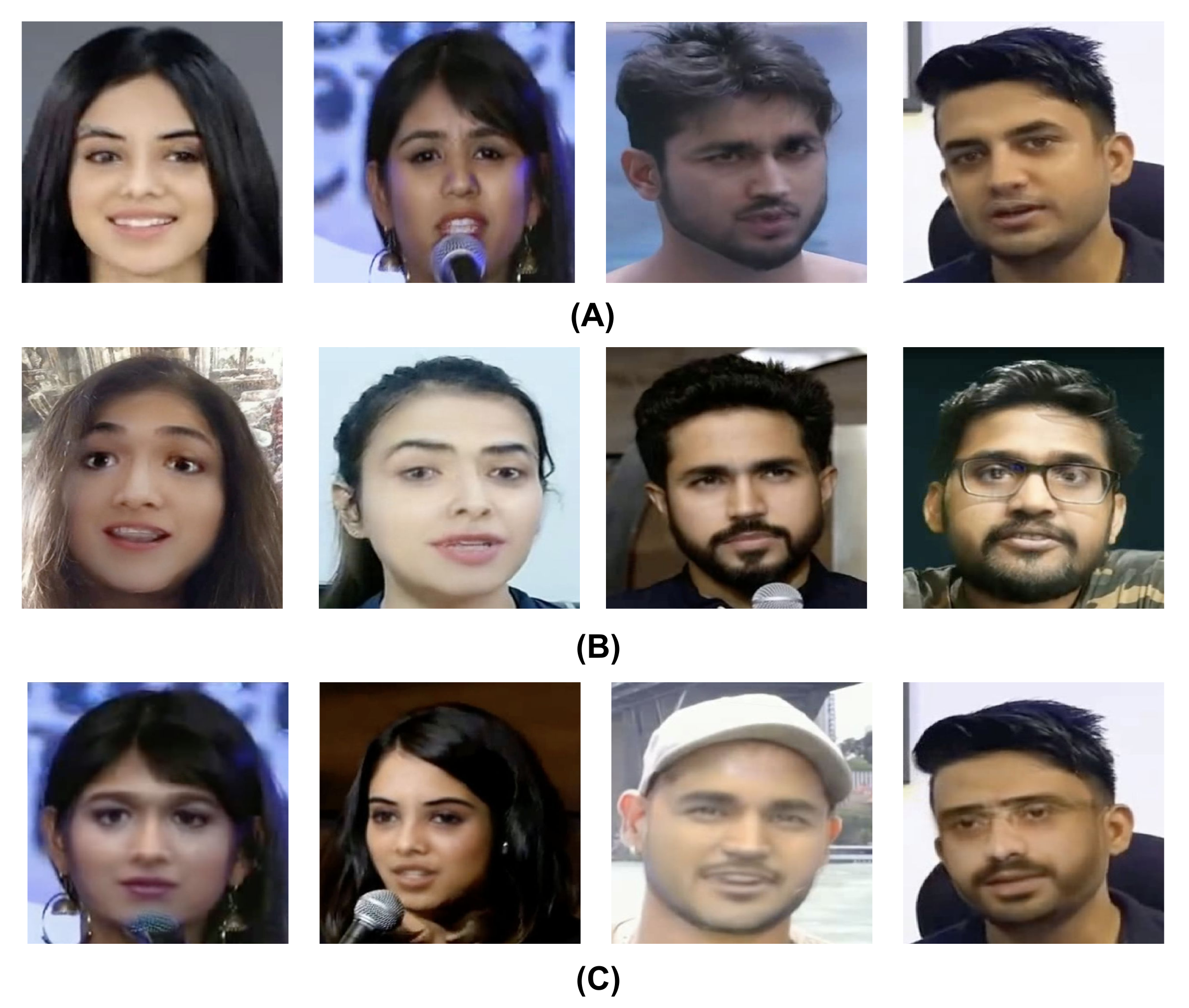}
\caption{Samples from (a) Successsion 1 (b) Succession 2 (c) Succession 3 of the proposed DeePhy dataset.}
\label{fig:diag}
\end{figure}

\noindent \textbf{Dataset Statistics:} The DeePhy dataset comprises a total of $100$ real videos and $5040$ deepfake videos generated with several succession of face swapping, with variations in the generation technique in each succession. Several existing deepfake datasets consist of videos taken in controlled environment where the variation of different lightning conditions is not captured properly. Moreover, in the controlled environment settings, variation in background, pose and expressions are kept at minimum which is contrary to real-life scenarios. In DeePhy dataset, the real videos of subjects are taken from \textit{Youtube} which is a publicly accessible platform with diverse distribution in gender, orientation, skin tone, size of face (in pixels), lighting conditions, background and presence of occlusion\footnote{The research is approved by the institutional ethical review committee.}. The dataset includes samples from various sub-ethnic groups. Some samples of the DeePhy dataset are shown in Figure~\ref{fig:diag}.

\begin{table}[t]
\centering
\caption{The generation techniques used in different successions of the DeePhy dataset. Successions 2 and 3 correspond to deepfake techniques applied two and three times in a row, respectively.}
\begin{tabular}{|c|c|c|}
\hline
\textbf{Succession 1}         & \textbf{Succession 2} & \textbf{Succession 3} \\ \hline
\multirow{3}{*}{FSGAN}       & FaceSwap             & FaceShifter          \\ \cline{2-3} 
                             & FaceShifter          & FaceSwap             \\ \cline{2-3} 
                             & FSGAN                &                      \\ \hline
\multirow{3}{*}{FaceSwap}    & FSGAN                & FaceShifter          \\ \cline{2-3} 
                             & FaceShifter          & FSGAN                \\ \cline{2-3} 
                             & FaceSwap             &                      \\ \hline
\multirow{3}{*}{FaceShifter} & FSGAN                & FaceSwap             \\ \cline{2-3} 
                             & FaceSwap             & FSGAN                \\ \cline{2-3} 
                             & FaceShifter          &                      \\ \hline
\end{tabular}
\label{datasetTable}
\end{table}

\noindent \textbf{Size and Format:} The total raw size of the dataset is around $30$ GB. The average duration of the videos is $\sim 20$ seconds and all videos are in $720p$ resolution with $25$ frames per second. The videos are in MPEG4.0 format.

\begin{figure}[] 
\centering
\includegraphics[scale=0.33]{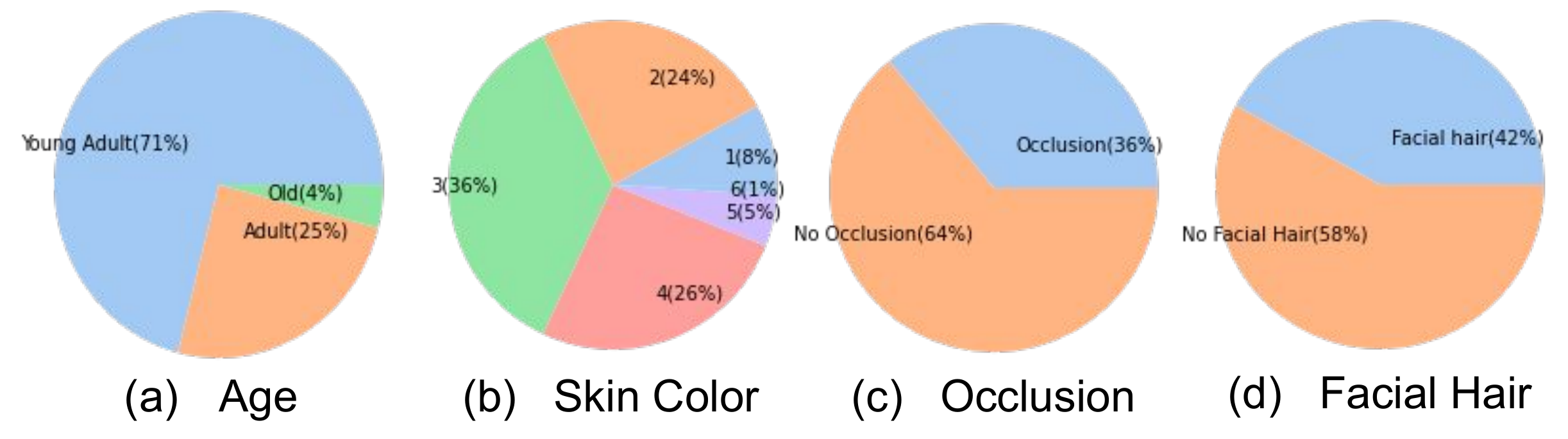}
\caption{Distribution of real videos in the proposed DeePhy dataset across multiple attributes: (a) Age (Young Adult/Adult/Old) (b) Skin Color (1/2/3/4/5/6) (c) Occlusion (Spectacles/Mic/Hair/Cap/Turban/Hijab/Scarf) and (d) Facial Hair (5oClockshadow/Moustache/Beard)}
\label{fig:stats}
\end{figure}

\noindent \textbf{Annotation and Diversity: } The proposed DeePhy dataset is annotated with $10$ attributes - Gender, Age, Skin Color, 5oClockshadow, Beard, Moustache, Spectacles, Shades, Mic, Cap/Turban/Hijab/Scarf and Hair Occlusion.
Gender is annotated as ``Male" or ``Female". Age is divided into three categories, people with apparent age between 18 (inclusive) to 30 belong to ``Young Adult", with apparent age between 30 (inclusive) to 55 belong to ``Adult"  and with apparent age greater than 55 (inclusive) belong to ``Old". The Skin Color annotations vary from $1$ to $6$ which corresponds to the $6$ skin color types in the Fitzpatrick scale. All the other attributes are binary and their values can either be ``Y" or ``N" which represents presence of the attribute and absence of the attribute, respectively. Figure~\ref{fig:stats} shows the diversity in attributes of the subjects in real videos.

\subsection{Synthesis Methods}
\label{SynthesisMethod}

We employ three synthesis methods namely, FSGAN~\cite{nirkin2019fsgan}, FaceSwap~\cite{faceswap}, and FaceShifter~\cite{faceshifter}. \\
\noindent\textbf{FSGAN}~\cite{nirkin2019fsgan} is a GAN-based deepfake generation technique used to perform face swapping of a source face onto a target video, which accounts for both pose and expression variations. FSGAN uses four different generators for each of the process of reenactment, face segmentation, inpainting, and Poisson blending. It applies an adversarial loss which utilizes a multi-scale discriminator consisting of multiple discriminators, each one operating on a different image resolution. For the proposed dataset, we use generator weights trained on data described in~\cite{nirkin2019fsgan}, and perform finetuning on the reenactment generator for each source video.
\\
\noindent\textbf{FaceSwap}~\cite{faceswap} is a deepfake generation technique based on autoencoders. It uses two autoencoders with a shared encoder that are trained to reconstruct training images of the source and the target face, respectively. The autoencoder output is blended with the rest of the image using Poisson image editing~\cite{10.1145/882262.882269}.  For DeePhy dataset, we train the autoencoder for each pair of source and target video. 
\\
\noindent\textbf{FaceShifter}~\cite{faceshifter} is a two-stage deepfake generation framework which was proposed for occlusion aware face-swapping with high fidelity. FaceShifter develops high-fidelity swapped faces by undertaking thorough blending of facial features, unlike prior face-swapping techniques that only used restricted information from target photos to synthesize faces. It transmits localised feature maps between faces by employing a new Adaptive Attentional Denormalization (AAD) layer. FaceShifter, unlike FSGAN, decreases the number of iterations by managing occlusions with a refinement network trained to examine the difference between the original and rebuilt image. For DeePhy dataset, the generator weights are used to train on data described in~\cite{faceshifter}, and inferencing is performed for each source-target pair.

\section{Experimental Framework and Baseline Results}
\label{sec:expt}

The experiments are performed on the DeePhy dataset. The dataset contains phylogenetic deepfakes generated by a combination of deepfake generation techniques. These generation techniques include FSGAN, FaceSwap, and FaceShifter. The performance of six deep learning models has been evaluated under different experimental settings. The dataset and evaluation protocols are described below.

\subsection{Dataset Protocol}
\label{DatasetProtocol}
The dataset is divided into a train set and a test set with an approximate 50-50 split. For this, $10$ frames are extracted per fake video. The total number of fake frames obtained are $50,400$. 
The train and test set comprises of approximately $27,100$ and $23,300$ frames respectively.
The split of the train and test set is made in a subject-disjoint manner such that a specific source/target belongs to one of the two sets. The division of frames also ensures that the three generation techniques \textit{FSGAN}, \textit{FaceSwap}, and \textit{FaceShifter} used to make phylogenetic deepfakes are equally represented in both the sets.

\begin{figure*}[t] 
\centering
\includegraphics[width=\textwidth]{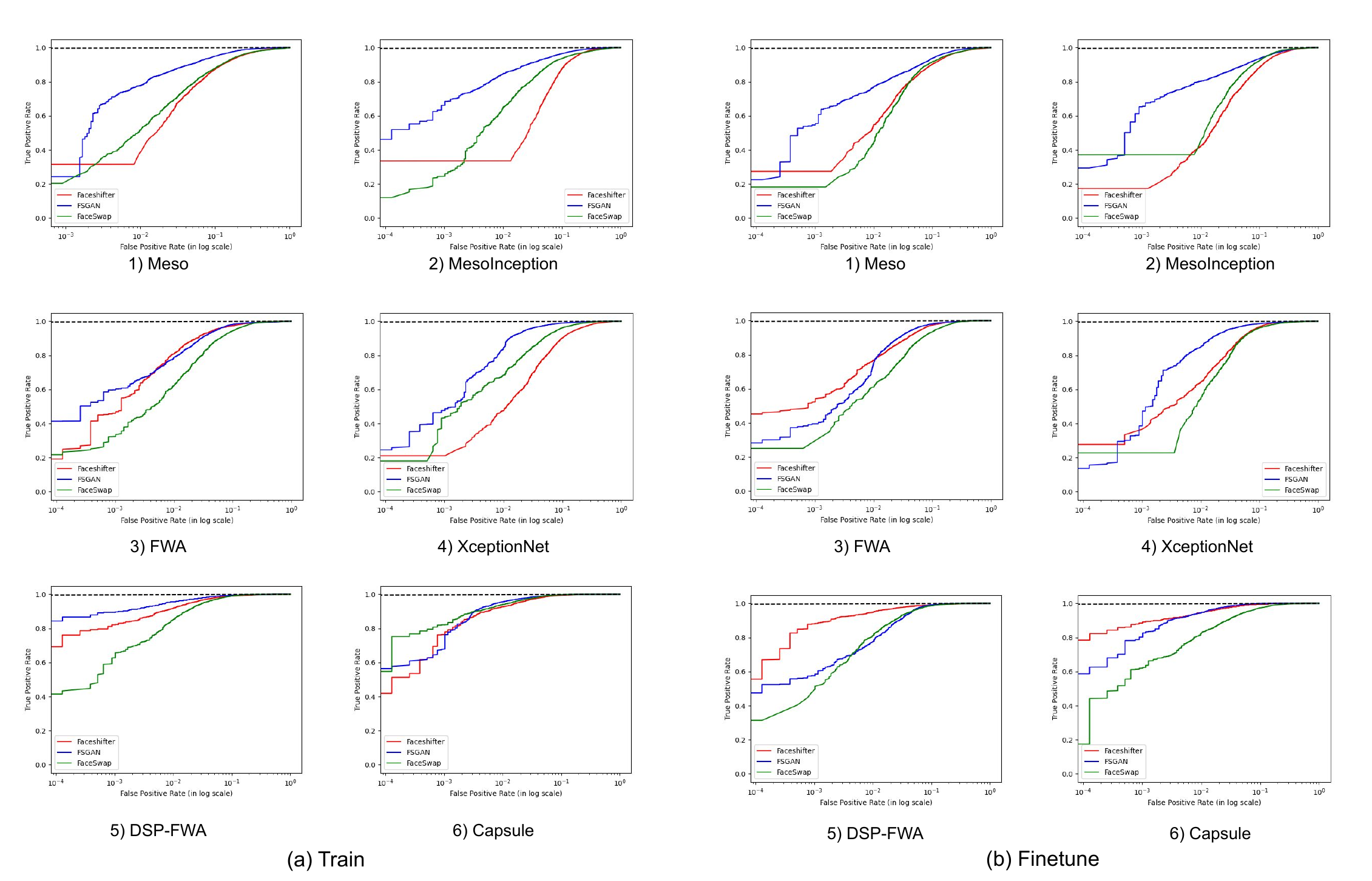}
\caption{The ROC curves (in log scale) corresponding to six state-of-the-art baseline models for the multi-label classification experiments involving Faceshifter, FSGAN, and FaceSwap (Protocol 2).}
\label{fig:roc}
\end{figure*}

\subsection{Evaluation Protocol}
\noindent We evaluate the performance of various state-of-the-art deepfake detection models on the test set of the DeePhy dataset under three evaluation protocols.

\noindent \textbf{Protocol 1: } This protocol pertains to detecting deepfakes and depicts the performance of state-of-the-art deepfake detection models on the proposed dataset. This also helps to gauge the quality of DeePhy dataset compared to other existing deepfake databases. In this experiment, the train-test split is approximately 50-50 and is performed in a subject disjoint manner to ensure that a particular source belongs to one of the two sets. The train set comprises of $48,576$ ($21,456$ real and $27,120$ fake) frames and the test set consists of $54,974$ ($31,694$ real and $23,280$ fake) frames. We train six deepfake detection models in a binary-classification setting to classify an input data sample as \textit{"real"} or \textit{"fake"}. We report the class-wise accuracy as well as the overall accuracy for the baseline models on the test set of this experiment.

\noindent \textbf{Protocol 2:} This protocol estimates the model's potential to identify the presence of different deepfake generation techniques. This protocol is a model attribution task where the trained model has to identify whether a particular generative algorithm was present in the sequence of algorithms that generated the deepfake. We employ the train set of DeepPhy to train a model in a multi-label classification setting.
We benchmark the performance of different models by their capability to correctly predict all the deepfake generation techniques employed in creation of a phylogenetic deepfake. In the framework of this protocol, the order of the generation techniques utilized in the creation of the deepfake is not conveyed by its performance. We report the overall accuracy which signifies the model's ability to correctly identify all the generation techniques involved irrespective of their sequence.

\noindent \textbf{Protocol 3:} The final protocol is a challenging model attribution task wherein the model needs to predict the sequence in which the generative techniques were employed to create a particular deepfake. We train a model in a multi-class classification setting to predict the sequence in which the deepfake generation techniques are used to create a phylogenetic deepfake. There are 18 possible sequences that can be employed to create deepfakes using the techniques discussed in Section \ref{SynthesisMethod}. We report the accuracy for each of the sequences as well as the overall accuracy for the baseline models. The models are trained in a multi-class classification setting.

In Protocol-2 and Protocol-3 described above, we conduct two experiments. In the first experiment, we train the model on the train set of the proposed dataset and report the performance on the test set (referred as \textit{Train} experiment). For the second experiment, we utilize the existing datasets containing deepfake samples generated using FSGAN, FaceSwap and FaceShifter techniques to pre-train a model (referred as \textit{Finetune} experiment). For this, we use the FaceForensics++ dataset~\cite{rossler2019faceforensics++} to collect samples of FaceSwap and FaceShifter generated deepfakes. The FSGAN deepfakes are generated using raw videos collected from CelebDF \cite{li2020celeb} dataset. The pre-training of the model using these samples is performed in a multi-label classification setting. We then finetune the model on the train set of our proposed dataset and report the performance.

\subsection{Models and Evaluation Metrics}
\label{EvaluationModels}
\noindent Six models are used to benchmark the proposed DeePhy dataset. For benchmarking, the following models are trained for identification of generation technique and prediction of sequence their sequence of generation.\\  
\noindent \textbf{XceptionNet}~\cite{chollet2017xception} is an acronym from extreme InceptionNet and is a neural network with separable convolutions with residual connections.\\
\noindent \textbf{MesoNet}~\cite{afchar2018mesonet} is a CNN-based deepfake detection method. There are two variants of MesoNet, namely, \textit{Meso4} and \textit{MesoInception4}. We use both Meso4 and Inception modules-based~\cite{szegedy2015going} MesoInception4 for our experiments. \\
\noindent \textbf{FWA}~\cite{li2018exposing} is a deepfake detection model with ResNet-50~\cite{he2016deep} as the underlying architecture. It targets artifacts and uses it as distinctive feature to distinguish between real and deepfake videos. The artifacts are caused due to affine face warping transformations. \\
\noindent \textbf{DSP-FWA} \cite{li2019exposing} is an improved technique based on FWA. It employs dual spatial pyramid strategy which includes a spatial pyramid pooling (SPP) module, on both image and feature level that better handles multi-scale issues.\\
\noindent \textbf{CapsuleNet}~\cite{nguyen2019use} uses capsule structures~\cite{sabour2017dynamic} based on VGG19~\cite{simonyan2014very} network as the underlying architecture. It takes into consideration spatial relationships of features in an image.

\noindent \textbf{Evaluation Metrics: } For the evaluation of first protocol, we report the classwise (\textit{"real"}, \textit{"fake"}) accuracy along with the overall accuracy for the deepfake detection task. In the evaluation of second protocol, we evaluate the overall multi-label classification performance of various detection models and compare each generation technique employed using ROC curves. Overall accuracy is the number of correctly classified instances in which all the generation techniques are correctly identified. It does not include cases where only a subset of the employed generation techniques are correctly predicted. Finally, in the third protocol, we evaluate the classwise and overall accuracy of the model in a multi-class classification setting wherein we consider each unique sequence as a separate class.

\begin{table}[]
\centering
\caption{Results for deepfake Detection experiment (Protocol 1) on the DeePhy dataset. Overall accuracy along with class-wise accuracy (\textit{"Real"} and \textit{"Fake"}) is reported.}
\begin{tabular}{|l|ccc|}
\hline
\multicolumn{1}{|c|}{\multirow{2}{*}{\textbf{Models}}} & \multicolumn{3}{c|}{\textbf{Accuracy}}                                                                          \\ \cline{2-4} 
\multicolumn{1}{|c|}{}                                 & \multicolumn{1}{l|}{\textbf{Real}} & \multicolumn{1}{l|}{\textbf{Fake}} & \multicolumn{1}{l|}{\textbf{Overall}} \\ \hline
MesoNet \cite{afchar2018mesonet}                                               & \multicolumn{1}{c|}{98.02}         & \multicolumn{1}{c|}{74.40}         & 84.85                                 \\ \hline
Meso-Inception \cite{afchar2018mesonet}                                        & \multicolumn{1}{c|}{97.84}         & \multicolumn{1}{c|}{75.72}         & 85.76                                 \\ \hline
FWA  \cite{li2018exposing}                                                  & \multicolumn{1}{c|}{98.86}         & \multicolumn{1}{c|}{71.63}         & 82.94                                 \\ \hline
XceptionNet \cite{chollet2017xception}                                           & \multicolumn{1}{c|}{99.24}         & \multicolumn{1}{c|}{72.54}         & 83.77                                 \\ \hline
DSP-FWA \cite{he2015spatial}                                               & \multicolumn{1}{c|}{99.70}         & \multicolumn{1}{c|}{78.25}         & 88.13                                 \\ \hline
Capsule  \cite{nguyen2019use}                                              & \multicolumn{1}{c|}{99.89}         & \multicolumn{1}{c|}{79.26}         & \textbf{88.89}                                 \\ \hline
\end{tabular}
\label{tab:pro1}
\end{table}

\begin{table}[]
\centering
\caption{Overall accuracy achieved on Protocol 2 using various baseline models.}
\begin{tabular}{|l|c|c|}
\hline
\textbf{Models} & \textbf{Train} & \textbf{Finetune} \\ \hline
MesoNet \cite{afchar2018mesonet}        & 73.18          & 74.23                         \\ \hline
Meso-Inception \cite{afchar2018mesonet} & 75.87          & 75.44                         \\ \hline
FWA \cite{li2018exposing}            & 82.78          & 82.11                         \\ \hline
XceptionNet \cite{chollet2017xception}    & 81.48          & 83.10                         \\ \hline
DSP-FWA \cite{he2015spatial}        & 88.77          &85.93                       \\ \hline
Capsule \cite{nguyen2019use}        &\textbf{90.86  }         &\textbf{87.97}                        \\ \hline
\end{tabular}
\label{tab:pro2}
\end{table}

\noindent \textbf{Implementation Details: } The source videos for the dataset were collected from YouTube and contain subjects of Indian origin. The videos were generated using FaceSwap \cite{faceswap}, FaceShifter \cite{faceshifter} and FSGAN \cite{nirkin2019fsgan} techniques using their published codes. For FaceSwap, each video was generated after 8 hours of training on two Nvidia DGX A100 system consisting of 8 GPUs of 80GB memory each. Similarly, videos were generated by employing pre-trained weights of FaceShifter using default parameters on two Nvidia RTX 3090 GPUs of 24 GB memory each. Furthermore, we fine-tuned the re-enactment generator of FSGAN for each source video and performed inferencing using Nvidia DGX A40 consisting of 48 GB memory. The dataset generation was completed in over 1100 hours with parallel use of the above mentioned GPUs. The baseline experiments of the proposed dataset are performed on Nvidia DGX station with four Tesla V-100 GPUs consisting of 32 GB memory each. The frames from videos are extracted using DSFD \cite{li2019dsfd}. While training for each protocol, the models are trained for 30 epochs with early stopping using Adam Optimizer having a learning rate of 0.0001 and keeping rest of the parameters as default.

\begin{table*}[h]
\centering
\scriptsize
\caption{Results for Protocol 3 for the DeePhy dataset. Each row depicts a different sequence of deepfake generation techniques. The overall accuracy and the accuracy corresponding to each sequence is reported for six models trained for both experiments. Each entry in Sequence column represents the order followed in phylogeny of the deepfake. S1 - FSGAN, S2 - FaceSwap, S3 - FaceShifter}
\begin{tabular}{|l|c|c|c|c|c|c|c|c|c|c|c|c|} 
\hline
\multirow{2}{*}{\textbf{Sequence}}                                            & \multicolumn{6}{c|}{\textbf{Training}}                                                                        & \multicolumn{6}{c|}{\textbf{Finetune}}                                                                        \\ 
\cline{2-13}
                                                                              & MesoNet & \begin{tabular}[c]{@{}c@{}}Meso-\\ Inception\end{tabular} & FWA   & XceptionNet & DSP-FWA & Capsule & MesoNet & \begin{tabular}[c]{@{}c@{}}Meso-\\ Inception\end{tabular} & FWA   & XceptionNet & DSP-FWA & Capsule  \\ 
\hline
S1                                                                            & 63.04   & 75.67                                                     & 68.60 & 68.45       & 69.39   & 61.70   & 57.14   & 62.94                                                     & 64.18 & 58.34       & 80.84   & 81.83    \\
S2                                                                            & 59.53   & 71.85                                                     & 61.59 & 59.76       & 69.56   & 56.08   & 63.04   & 63.27                                                     & 75.71 & 47.17       & 74.63   & 60.60    \\
S3                                                                            & 85.38   & 86.16                                                     & 78.76 & 71.50       & 91.59   & 74.10   & 71.20   & 76.26                                                     & 85.09 & 74.20       & 91.89   & 94.30    \\
S1-S2                                                                         & 40.22   & 49.79                                                     & 60.00 & 43.04       & 61.33   & 67.39   & 44.72   & 57.74                                                     & 51.68 & 46.11       & 51.41   & 84.14    \\
S1-S3                                                                         & 41.85   & 35.17                                                     & 46.41 & 40.68       & 57.26   & 74.47   & 47.47   & 35.76                                                     & 41.95 & 52.89       & 71.33   & 43.66    \\
S1-S1                                                                         & 45.11   & 56.32                                                     & 75.16 & 65.50       & 81.08   & 71.45   & 43.10   & 70.10                                                     & 67.83 & 69.38       & 80.20   & 47.74    \\
S2-S1                                                                         & 45.55   & 51.23                                                     & 64.39 & 63.90       & 89.01   & 53.53   & 44.18   & 45.73                                                     & 50.66 & 40.22       & 73.65   & 73.97    \\
S2-S3                                                                         & 54.28   & 59.29                                                     & 68.22 & 45.39       & 74.26   & 69.70   & 54.50   & 52.93                                                     & 55.72 & 53.87       & 76.44   & 74.57    \\
S2-S2                                                                         & 64.97   & 61.83                                                     & 62.78 & 54.82       & 64.52   & 42.20   & 55.00   & 58.62                                                     & 58.01 & 64.13       & 70.88   & 68.15    \\
S3-S1                                                                         & 32.22   & 39.68                                                     & 46.41 & 44.30       & 70.21   & 80.38   & 56.86   & 45.98                                                     & 43.01 & 40.80       & 71.86   & 54.87    \\
S3-S2                                                                         & 58.27   & 56.17                                                     & 61.12 & 49.49       & 74.04   & 55.51   & 56.63   & 45.74                                                     & 54.44 & 59.91       & 82.95   & 91.19    \\
S3-S3                                                                         & 91.90   & 79.29                                                     & 83.20 & 76.68       & 85.45   & 95.00   & 90.66   & 82.10                                                     & 84.50 & 98.18       & 98.79   & 74.34    \\
S1-S2-S3                                                                      & 39.11   & 39.67                                                     & 38.69 & 44.61       & 72.49   & 80.94   & 37.99   & 52.80                                                     & 43.09 & 39.60       & 55.39   & 55.73    \\
S1-S3-S2                                                                      & 26.08   & 32.50                                                     & 35.91 & 28.79       & 30.76   & 47.68   & 30.14   & 30.68                                                     & 25.78 & 32.73       & 50.16   & 57.19    \\
S2-S1-S3                                                                      & 21.64   & 30.11                                                     & 25.64 & 30.39       & 37.69   & 71.79   & 31.55   & 25.27                                                     & 19.75 & 21.30       & 53.05   & 55.30    \\
S2-S3-S1                                                                      & 28.71   & 28.53                                                     & 34.74 & 34.67       & 48.84   & 52.73   & 28.47   & 26.72                                                     & 23.92 & 31.65       & 56.52   & 38.10    \\
S3-S1-S2                                                                      & 30.92   & 31.80                                                     & 31.13 & 37.96       & 29.82   & 49.63   & 41.88   & 35.03                                                     & 35.16 & 53.36       & 52.04   & 50.78    \\
S3-S2-S1                                                                      & 25.30   & 34.50                                                     & 35.74 & 41.47       & 36.08   & 61.73   & 34.84   & 30.14                                                     & 29.51 & 33.92       & 63.92   & 33.44    \\ 
\hline
\textbf{\begin{tabular}[c]{@{}l@{}}Overall \\ Accuracy\end{tabular}} & 45.15   & 49.86                                                     & 54.04 & 50.77       & 60.65   & 62.87   & 48.96   & 48.92                                                     & 49.16 & 49.61       & 69.38   & 60.15    \\
\hline
\end{tabular}
\label{tab:protocol3}
\end{table*}

\subsection{Benchmark Results}
\label{sec:results}
\noindent Using DeePhy dataset in conjunction with other existing deepfake datasets, we benchmark the performance on the proposed dataset using existing and state-of-the-art techniques. As mentioned in section \ref{sec:expt}, the baseline experiments are performed for three protocols aimed at deepfake detection, identification of deepfake generation techniques and predicting the sequence of their occurrence. The experiments of Protocol-1 are performed in a binary-classification setting whereas for Protocol-2, it is a multi-label classification setting, and Protocol-3 is performed in a multi-class classification setting.

\noindent \textbf{Protocol 1: }The class-wise and overall accuracy achieved by different state-of-the-art models on deepfake detection task (classification between real and fake media) which is reported in Table \ref{tab:pro1}. The recent models show a better class-wise accuracy of \textit{"real"} samples as compared to the \textit{"fake"} samples. This shows that the existing models lack the capability to detect phylogenetic deepfakes. The best overall accuracy is achieved by CapsuleNet i.e. 88.89\% followed by DSP-FWA and Meso-Inception at 88.13\% and 83.77\% respectively. The detection accuracy of the baseline models when compared to other existing deepfake datasets ~\cite{wilddf, deepfaketimit, celebdf, FaceForensics, ff++} reflects the quality of generated deepfakes in the proposed DeePhy dataset. We believe that a sub-set (Succession 1 deepfakes) of the proposed DeePhy dataset can also be used with an inclusion in the existing datasets for more robust training of the models. This will also ensure proper representation of Indian origin which is currently under-represented in the existing deepfake datasets.

\noindent \textbf{Protocol 2: }The results of Protocol-2 as shown in Table \ref{tab:pro2} measures the model’s potential to identify the presence of different deepfake generation techniques. It is observed that the model is able to identify the samples generated from FSGAN better when compared to other generation techniques. This is because FSGAN has the least training time amongst the three deepfake generation models and is not an occlusion-aware technique. This is further corroborated by Figure \ref{fig:roc} which shows the frame-level ROC curves obtained on the proposed DeePhy dataset. CapsuleNet outperforms all other techniques with a significant margin for Train experiment achieving an overall accuracy of 90.86\%. CapsuleNet performs the best for the Finetuning experiment as well achieving a accuracy of 87.97\%, outperforming DSP-FWA by a small margin. CapsuleNet and DSP-FWA are followed by XceptionNet achieving 81.48\% and 83.10\% for Train and Finetune experiment, respectively. CapsuleNet achieves the highest class-wise AUC score followed by DSP-FWA and XceptionNet. The highest performing model provides an accuracy of approximately 90\%, yielding a large scope for improvement in deepfake detection techniques in a multi-label setting. 

\noindent \textbf{Protocol 3:} Protocol-3 is a phylogeny sequence prediction task wherein the sequence of different generative techniques is predicted. The performance achieved by the different models is listed in Table \ref{tab:protocol3}. The highest overall accuracy is 62.87\% achieved by the Capsule Network. However, most models achieve an overall accuracy less than 50\%  with or without fine-tuning. It can be observed that as the number of successions in the sequence increases the accuracy of sequence prediction decreases. This is because phylogeny of a deepfake increases with successions. The low performance achieved by the models provides an opportunity to improve the performance of deepfake generation model sequence prediction. \\

\section{Discussion and Future Work}
\noindent For the first time in the literature, this research presents a new research thread of deepfake phylogeny and its detection. We propose DeePhy, a novel Deepfake Phylogeny dataset consisting of 5040 deepfake videos generated using multiple generation techniques sequentially for upto three successions. \textit{The DeePhy dataset poses the challenge of model attribution and identification of the sequence of generation techniques used while creating a deepfake image or video.} The proposed dataset is benchmarked using six model architectures, popularly used in deepfake detection techniques. On performing various experiments, we observe the following:
\setlist{nolistsep}
\begin{itemize}[noitemsep]
  \item With easily available deepfake generation techniques, deepfakes can be created using multiple operations applied sequentially. The average accuracy obtained by current deepfake detection methods to identify the sequence of face manipulations used to generate deepfakes is 54.36\%, which shows that deepfake detection algorithms are not ready to address phylogenetic deepfakes and more research is needed in the field of Deepfake Phylogeny.
  \item The task of model attribution has several real-life applications but the performance of detection models for this task is not up to the mark.
  \item The existing deepfake datasets~\cite{wilddf, deepfaketimit, celebdf, FaceForensics, ff++} under-represent subjects of Indian origin. Succession 1 samples of DeePhy dataset can be used in inclusion with existing datasets to fill this void and make the deepfake models more robust and generalised over all ethnicities/origins. 
\end{itemize}
The existing research on synthetic media concerns with association of binary label - \textit{"real"} and \textit{"fake"} to an image. In future, more efforts are required towards deepfake detection where along with assigning labels (\textit{"real"} and \textit{"fake"}), model attribution and sequence of generation techniques employed should also be studied. As discussed in earlier, this additional information of a deepfake image/video has a lot of potential and will facilitate advancements in real-life scenarios of plagiarism detection, forgery detection, and reverse engineering of deepfakes. 
\section{Acknowledgement}
\noindent This research is supported through a grant from Ministry of Home Affairs, Government of India. K. Thakral is partially supported by the PMRF Fellowship. S. Mittal is partially supported by the UGC-Net JRF Fellowship.  M. Vatsa is partially supported through the Swarnajayanti Fellowship. 

{\small
\bibliographystyle{ieeetr}
\bibliography{egbib}
}

\end{document}